\RequirePackage{fix-cm}
\documentclass[final]{svjour3}
\smartqed

\usepackage{microtype}
\usepackage{graphicx}
\usepackage{subfigure}
\usepackage{booktabs} 
\usepackage{amsfonts}
\usepackage{amsmath}
\usepackage{epsf}
\usepackage{units}
\usepackage{slashed}
\usepackage{MnSymbol}
\usepackage{natbib}
\usepackage{color}

\usepackage{hyperref}

\renewcommand{\vec}[1]{{\mathbf #1}}
\newcommand{\vecg}[1]{{\pmb #1}}

\DeclareMathOperator{\diag}{diag}
\DeclareMathOperator{\Cov}{Cov}
\DeclareMathOperator{\Corr}{Corr}
\DeclareMathOperator{\Var}{Var}

\begin{document}
\title{Scrutinizing XAI using linear ground-truth data with suppressor variables}
\titlerunning{Scrutinizing XAI with linear ground-truth data}  

\author{Rick Wilming \and
        Céline Budding \and
        Klaus-Robert Müller \and
        Stefan Haufe
}

\institute{ Rick Wilming \at
            Technische Universität Berlin, Germany\\
            \email{rick.wilming@tu-berlin.de}
            \and
              Céline Budding \at
              Eindhoven University of Technology\\
              \email{c.e.budding@tue.nl}           
            \and
              Klaus-Robert Müller \at
              Technische Universität Berlin, Germany\\
              BIFOLD -- Berlin Institute for the Foundations of Learning and Data, Berlin, Germany\\
              Korea University, Seoul, South Korea\\
              Max Planck Institute for Informatics, Saarbr{\"u}cken, Germany\\
              \email{klaus-robert.mueller@tu-berlin.de}           
           \and
           Stefan Haufe \at
              Technische Universität Berlin, Germany
              \at
              Physikalisch-Technische Bundesanstalt Berlin, Germany
              \at
              Charit{\'e} -- Universitätsmedizin Berlin, Germany\\
              \email{haufe@tu-berlin.de}
}

\date{June 2023}

\maketitle

\begin{abstract}
Machine learning (ML) is increasingly often used to inform high-stakes decisions. As complex ML models (e.g., deep neural networks) are often considered black boxes, a wealth of procedures has been developed to shed light on their inner workings and the ways in which their predictions come about, defining the field of `explainable AI' (XAI). Saliency methods rank input features according to some measure of `importance’. Such methods are difficult to validate since a formal definition of feature importance is, thus far, lacking. It has been demonstrated that some saliency methods can highlight features that have no statistical association with the prediction target (suppressor variables). To avoid misinterpretations due to such behavior, we propose the actual presence of such an association as a necessary condition and objective preliminary definition for feature importance. We carefully crafted a ground-truth dataset in which all statistical dependencies are well-defined and linear, serving as a benchmark to study the problem of suppressor variables. We evaluate common explanation methods including LRP, DTD, PatternNet, PatternAttribution, LIME, Anchors, SHAP, and permutation-based methods with respect to our objective definition. We show that most of these  methods are unable to distinguish important features from suppressors in this setting. 
\end{abstract}


\section{Declarations}
\vspace{-0.15cm}
\textbf{Funding}---This result is part of a project that has received funding from the European Research Council (ERC) under the European Union’s Horizon 2020 research and innovation programme (Grant agreement No. 758985).
KRM also acknowledges support by the German Ministry for Education and Research as BIFOLD -- Berlin Institute for the Foundations of Learning and Data (ref.\ 01IS18025A and ref.\ 01IS18037A), and the German Research Foundation (DFG) as Math+: Berlin Mathematics Research Center  (EXC 2046/1, project-ID: 390685689),
Institute of Information \& Communications Technology Planning \& Evaluation (IITP) grants funded by the Korea Government (No. 2019-0-00079,  Artificial Intelligence Graduate School Program, Korea University).\\

\vspace{-0.15cm}
\noindent\textbf{Conflicts of interest/Competing interests}---The authors declare no conflicts of interest/competing interests.\\

\vspace{-0.15cm}
\noindent\textbf{Availability of data and material}---All data used here can be generated using the provided code.\\

\vspace{-0.15cm}
\noindent\textbf{Code availability}---\url{https://github.com/braindatalab/scrutinizing-xai}\\



\vspace{-0.15cm}
\noindent \textbf{Ethics approval}---Not applicable.

\noindent \textbf{Consent to participate}---Not applicable.

\noindent \textbf{Consent for publication}---Not applicable.


\vspace{-0.4cm}
\section[]{Introduction}\label{sec:introduction}
\vspace{-0.15cm}
With AlexNet~\citep{krizhevsky_imagenet_2012} winning the ImageNet
competition, the machine learning (ML) community started into a new era. Within few years, novel models achieved massive leaps in performance for challenging problems in computer vision, natural language processing, and reinforcement learning \citep[e.g., ][]{silver_mastering_2017, lecun2015deep, jaderberg2015spatial}. In several real-world tasks, ML models became on par with human experts or achieved even super-human performance \citep{silver_mastering_2017}. Nowadays, there are increasing efforts to also leverage their predictive power in fields such as healthcare and criminal justice, where they may support high-stake decisions that have a profound impact on human lives~\citep{rudin_stop_2019,lapuschkin2019unmasking}. 

The complexity of current ML models makes it hard for humans to understand the ways in which their predictions come about. Especially in highly safety- or otherwise critical fields such as medicine, finance, or automatic driving, ethical and legal considerations have led to the demand that predictions of ML models should be `transparent', establishing the field of `interpretable’ or `explainable' AI \citep[XAI, e.g.,][]{samek2019explainable,dombrowski2022towards}. Current XAI approaches can be categorized along various dimensions~\citep{arrieta_explainable_2019}. Some methods provide `explanations’ for single input examples (\emph{instance-based}), while others can be applied to entire models only (\emph{global}). A common paradigm is to provide `importance’ or `relevance’ scores for single input features. Respective methods are called `saliency' or `heat' mapping approaches. 
Another distinction is made between model-agnostic methods~\citep[e.g.][]{strumbelj_explaining_2014, ribeiro_why_2016, lundberg_unified_2017}, which are based on a model’s output only, and methods that are tailored to a specific class of models~\citep[e.g., neural networks, ][]{zeiler_visualizing_2014,springenberg_striving_2015,bach_pixel-wise_2015, binder_layer-wise_2016, montavon_explaining_2017,kim_interpretability_2018,montavon2018methods,samek2021explaining}. 
Finally, linear ML models with sufficiently few features as well as shallow decision trees have been considered intrinsically `interpretable’~\citep{rudin_stop_2019}, a notion that has also been challenged~\citep{haufe_interpretation_2014,lipton_mythos_2017,poursabzi2021manipulating}, and that is further scrutinized here.

It is understood that XAI methods can serve quality control purposes only under the provision of being trustworthy themselves. However, it is still under scientific debate what specific formal problems XAI is supposed to solve and what requirements respective methods should fulfill~\citep{lipton_mythos_2017, doshi-velez_towards_2017, murdoch_definitions_2019}. Existing formulations of such requirements are often relatively vague and lack precise mathematical language. Terms like `explainable' or `interpretable' are used by many XAI authors without specifying \emph{how} results of a given method should be interpreted, i.e., what exact formal conclusions can be deduced. Authors of XAI papers frequently suggest interpretations that are either not formally justified or not precise enough to be formally verified. LIME \citep{ribeiro_why_2016}, for example, includes the following example: ``A model predicts that a patient has the flu, and LIME highlights the symptoms in the patient’s history that led to the prediction. Sneeze and headache are portrayed as contributing to the `flu' prediction, while `no fatigue' is evidence against it. With these, a doctor can make an informed decision about whether to trust the model's prediction.'' As we will discuss, nescience about the capabilities of XAI methods can lead to misinterpretations in practice. 


Importantly, the lack of quantifiable formal criteria also currently prohibits the objective validation of XAI. Rather than using ground-truth data, current validation schemes are often either restricted to subjective qualitative assessments or use surrogate performance metrics such as the change in model output or performance when manipulating or omitting single features \citep[e.g.,][]{samek2016evaluating, fong2017interpretable, hooker_benchmark_2019, alvarez-melis_robustness_2018}. 
In this paper, we aim to make a first step towards an \emph{objective} validation of saliency methods. To this end, we devise a purely data-driven criterion of feature importance, which defines the superset of features that any XAI method may reasonably identify. Based on this definition, we generate simple synthetic ground-truth data with linear structure, which we use to quantitatively benchmark a multitude of existing XAI appproaches including LIME~\citep{ribeiro_why_2016}, SHAP~\citep{lundberg_unified_2017}, and LRP~\citep{bach_pixel-wise_2015} with respect to their \emph{explanation performance}.

\vspace{-0.4cm}
\section{Formalization of feature importance}\label{sec:formalization}
\vspace{-0.15cm}
Let us consider a supervised prediction task, where a model $f^{\boldsymbol{\theta}}: \mathbb{R}^D \rightarrow \mathcal{Y}$ learns a mapping from a $D$-dimensional feature space to a label space $\mathcal{Y}$ from a set of $N$ \textit{i.i.d.} training examples $\mathcal{D} = \{(\vec{x}^n, y^n)\}_{n=1}^N, \vec{x}^n \in \mathcal{F} \subseteq \mathbb{R}^D, y^n \in \mathcal{Y}, n \in \{1, \ldots, N\}$. The $\vec{x}^n$ and $y^n$ are realizations of random variables $\vec{X}$ and $Y$ with joint probability density function $p_{\vec{X},Y}(\vec{x}, y)$.
Saliency maps may either be obtained for entire ML models or on a single-instance basis. It is, thus, a function $\vec{s}(f^{\boldsymbol{\theta}}, \vec{x}^\ast, \mathcal{D}) \in \mathbb{R}^D$ that depends on the model $f^{\boldsymbol{\theta}}$ as well (optionally) the training data $\mathcal{D}$ and/or an input example $\vec{x}^\ast$. The map $\vec{s}$ is supposed to quantify the `importance' of each feature $d \in \{1, \ldots, D\}$ either for the prediction of the sample $\vec{x}^\ast$ or for the predictions of the model $f^{\boldsymbol{\theta}}$ in general, according to some criterion. Ideally, one would like to have a way of defining the `correct' saliency map for a certain combination of model and data. Coming up with such a definition is, however, difficult. We, therefore, constrain ourselves to the simpler problem of partitioning the set of features into `important' and `unimportant' ones. Thus, we are looking for functions $\vec{h}$, where $\vec{h}(f^{\boldsymbol{\theta}}, \vec{x}^\ast, \mathcal{D}) \in \{0, 1\}^D$. Here, $\mathcal{F}^+ := \{d : h_d(f^{\boldsymbol{\theta}}, \vec{x}^\ast, \mathcal{D}) = 1\}$ is the set of `important' features and $\mathcal{F}^- := \{d : h_d(f^{\boldsymbol{\theta}}, \vec{x}^\ast, \mathcal{D}) = 0\}$ is the set of `unimportant' features. For a given saliency map $\vec{s}$, a corresponding dichotomization function $\vec{h}$ can be obtained by tresholding its output, for example based on a statistical hypothesis test. 

\vspace{-0.4cm}
\subsection{Importance as influence on the model decision}
\vspace{-0.15cm}
The indicator function $\vec{h}$ facilitates possible formalizations of `importance'. Most current saliency methods -- implicitly or explicitly -- usually seek to identify those features that significantly influence the decision of a model. For some models, the corresponding sets, $\mathcal{F}_{\text{model}}^+$ and $\mathcal{F}_{\text{model}}^-$, can indeed be defined in a straightforward manner. Examples are linear models, for which $\mathcal{F}_{\text{model}}^{+}$ can be defined as the set of features with non-zero (not significantly different from zero) model coefficients. For more complex models, such a direct definition is, however, more difficult, and we refrain from attempting a more precise formalization here. 

\vspace{-0.4cm}
\subsection{Importance as statistical relation to the target}
\vspace{-0.15cm}
It is often implicitly assumed that XAI methods provide qualitative or even quantitative insight about statistical or mechanistic relationships between the input and output variables of a model~\citep{ribeiro_why_2016, binder_layer-wise_2016}. In other words, it is asserted that $\mathcal{F}^{+}$ must contain only features that are at least in some way structurally or statistically related to the prediction target. As an example, a brain region that is highlighted by an XAI method as `important’ for predicting a neurological disease will typically be interpreted as a correlate or even causal drive of that disease and be discussed as such. Such interpretations are, however, invalid, as it is possible that features lacking any structural or statistical relationship to the prediction target do significantly reduce the model's prediction error (thus, are in $\mathcal{F}_{\text{model}}^{+}$) \citep{haufe_interpretation_2014}. Such features have been termed suppressor variables \citep{conger_revised_1974,friedman2005graphical}.

Suppressor variables may contain side information, for example, on the correlation structure of the noise, that can be used by a model to predict better. But they themselves do not provide any satisfactory `explanation' about the actual relationship between input and output variables \citep{haufe_interpretation_2014}. Such features are prone to be misinterpreted, which could have severe consequences in high-stakes domains. We, therefore, argue that a genuine statistical dependency between a feature and the response variable should be a prerequisite for that feature to be considered important. In other words, the set of important features identified by any saliency method should be a subset of a set $\mathcal{F}_{\text{dep}}^{+}$ that can be defined based on the data alone as 
\begin{align}\label{eq:objective_importance}
\mathcal{F}_{\text{dep}}^{+} :=  \{d \; | \; X_{d} \,
\slashed{\upmodels} \, Y\} \;,
\end{align}
where $X_{d} \, \slashed{\upmodels} \, Y \Leftrightarrow p(x_d, y) \neq p(x_d) \cdot p(y)$ for some choice of $x_d, y$. The set of unimportant features is defined as the complement $\mathcal{F}_{\text{dep}}^{-} = \{d \; | \; X_{d} \,
\upmodels \, Y\} = \{1, \ldots, D\} - \mathcal{F}_{\text{dep}}^{+}$.

Notably, a data-driven mathematical definition of feature importance such as ours also provides a recipe to generate ground-truth reference data with known sets of important features. This paves the way for an objective evaluation of XAI methods, which is the purpose of this paper. 





\vspace{-0.4cm}
\section{Suppressor variables}\label{sec:supppressors}
\vspace{-0.15cm}
To understand how suppressor variables can cause misinterpretations for existing saliency methods, consider the following linear generative model \citep[c.f.,][]{haufe_interpretation_2014} with two features $x_1, x_2 \in \mathbb{R}$ and a response (target) variable $y \in \mathbb{R}$:
\begin{align}
    \label{eq:suppressor}
    x_1 &= a_1 \varsigma + d_1 \rho \\
\nonumber    x_2 & =  d_2 \rho \\
\nonumber    y  & = \varsigma \;.
\end{align}
%
%
Here, $\varsigma$ and $\rho \in \mathbb{R}$ are random variables called the \emph{signal} and \emph{distractor}, respectively, and $a_1, d_{1}$ and $d_2 \in \mathbb{R}$ are non-zero coefficients. The mixing weight vectors $\vec{a} = [a_1, 0]^\top$ and $\vec{d} = [d_1, d_2]^\top$ are called signal and distractor \emph{patterns}, respectively \citep[see][]{haufe_interpretation_2014,kindermans_learning_2017}. The learning task is to predict labels $y$ from features $\vec{x} = [x_1, x_2]^\top$. This task is solvable using $x_1$ alone, since $x_1$ and $y$ share the common signal $\varsigma$. However, the presence of the distractor in $x_1$ limits the achievable prediction accuracy. Since $x_2$ and $y$ do not share a common term, no prediction of $y$ above chance-level is possible using $x_2$. However, a bivariate model using both features can eliminate the distractor so that the label can be perfectly recovered. Specifically, the linear model $f^\vec{w}(\vec{x}) = \vec{w}^\top \vec{x}$ with demixing weight vector (also called extraction \emph{filter}) $\vec{w} = [\nicefrac{1}{a_1}, -\nicefrac{d_1}{(a_1 d_2)}]^\top$ achieves that:
\begin{align}
    \label{eq:weight-with-features}
    \vec{w}^\top \vec{x} = \varsigma + \frac{d_1}{a_1} \rho  - \frac{d_1}{a_1 d_2} d_2 \rho = \varsigma = y \;.
\end{align}
According to the terminology introduced above, the set of features statistically related to the target is $\mathcal{F}_{\text{dep}}^+ = \{1\}$, while the set of `influential' features is $\mathcal{F}_{\text{model}}^+ = \{1, 2\}$. The influence of $x_2$ on the prediction is certified by the non-zero coefficient $w_2 = -\nicefrac{d_1}{(a_1 d_2)}$ of the optimal prediction model. Depending on the coefficients $a_1$, $d_1$ and $d_2$, this influence can have positive or negative polarity, and its strength can be smaller or bigger than $w_1 = \nicefrac{1}{a_1}$, provoking diverse conjectures about the nature of its influence. However, $x_2$ itself has no statistical relationship to $y$ by construction. It is, therefore, a suppressor variable \citep{horst_prediction_1941,conger_revised_1974,friedman2005graphical,haufe_interpretation_2014}. 

In a real problem setting, $y$ could be a disease to be diagnosed and $\varsigma$ could be a (perfect) physiological marker. The baseline level of that marker could, however, be different for the two sexes, encoded in the distractor $\rho$, {even if the prevalance of the disease does not depend on sex}. A bivariate ML model can subtract the sex-specific baseline and, thereby, diagnose the disease more accurately than a univariate model based on the measured marker alone. A clinician confronted with the influence of sex on the model decision may thus erroneously conclude that sex is a factor that also correlates with or even causally influences the presence of the disease. Such examples refute the widely accepted notion \citep{ribeiro_why_2016, rudin_stop_2019} that linear models are easy to interpret \citep{haufe_interpretation_2014}.


\vspace{-0.4cm}
\section{Methods}\label{sec:classifiers}
\vspace{-0.15cm}
We generate synthetic images of two classes in the spirit of the linear example introduced in Section~\ref{sec:supppressors}; thus with known sets $\mathcal{F}_{\text{dep}}^+$ of class-specific pixels. These are then used to train linear classifiers to discriminate the two classes. The resulting models are analyzed by a multitude of XAI approaches in order to obtain `saliency' maps. The performance of these methods w.r.t. recovering (only) pixels in $\mathcal{F}_{\text{dep}}^+$ is then quantitatively assessed using appropriate metrics. The techniques used in these steps are described in the following.

\vspace{-0.4cm}
\subsection{Data generation}\label{subsec:data-generation}
\vspace{-0.15cm}
Following~\citet{haufe_interpretation_2014}, we extend the
two-dimensional example of a suppressor variable \eqref{eq:suppressor} and create synthetic data sets 
$\mathcal{D} = \{(\vec{x}^n, y^n)\}_{n=1}^N$ of \textit{i.i.d.}
observations $(\vec{x}^n \in \mathbb{R}^D, y^n \in \{-1, 1\})$ according to the generative model 
\begin{align}
    \label{eq:synthetic-data-model}
    \vec{x} = \lambda_1 \vec{a} y + \lambda_2 \vec{d} \rho + \lambda_3 \vecg{\eta},
\end{align}
with activation pattern $\vec{a} \in \mathbb{R}^{D}$ and distractor pattern $\vec{d} \in \mathbb{R}^{D}$. The signal is directly encoded using binary class labels $y \in \{-1, 1\} \sim \mathrm{Bernoulli}(\nicefrac{1}{2})$.
The distractor $\rho \in \mathbb{R} \sim \mathcal{N}(0, 1)$ is sampled from a standard normal distribution. The multivariate noise $\vecg{\eta} \in \mathbb{R}^{D} \sim \mathcal{N}(\vec{0}, \vecg{\Sigma})$ is modelled according to a $D$-dimensional multivariate Gaussian
distribution with zero mean and random covariance matrix $\vecg{\Sigma} = \vec{V} \vec{E} \vec{V}^\top$, where $\vec{V}$ is uniformly sampled from the space of orthogonal matrices and $\vec{E} = \diag(\vec{e})$ is a diagonal matrix of eigenvalues sampled as $\vec{e} = [e_1+c, \ldots, e_D+c]$, where $c:=\nicefrac{\max(\vec{e})}{100}$ and $e_d \sim \mathcal{U}(0, 1), d \in \{1, \ldots, D\}$. The signal, distractor, and noise \emph{components} $\vec{a} [y^1, \ldots, y^N]$, $\vec{d} [\rho^1, \ldots, \rho^N]$, and $[\vecg{\eta}^1, \ldots, \vecg{\eta}^N] \in \mathbb{R}^{D \times N}$ are normalized by their respective Frobenius norms, e.g., $[\vecg{\eta}^1, \ldots, \vecg{\eta}^N] \leftarrow [\vecg{\eta}^1, \ldots, \vecg{\eta}^N] / \sqrt{ \sum_{n=1}^N \sum_{d=1}^D |\eta_d^n|^2}$. The observed features are then obtained as a weighted sum of the three components, where the factors $\lambda_{i}$ with $\sum_{i}^{3} \lambda_{i} = 1$ adjust the influence of each component in the sum, defining the signal-to-noise ratio (SNR). For a given SNR factor $\lambda_1$, we here set $\lambda_2 = \lambda_3 = \nicefrac{(1-\lambda_1)}{2}$. 

Note that the activation pattern $\vec{a}$ represents the ground truth feature importance map according to our definition~\eqref{eq:objective_importance}. To be comprehensible by a human, it would be desirable for $\vec{a}$ to have a simple structure (e.g. sparse, compact). As it is an intrinsic property of the data, this cannot be ensured in practice, though. However, estimates of $\vec{a}$ can be biased towards having simple structure.

\vspace{-0.4cm}
\subsection{Classifiers}\label{subsec:classifiers}
\vspace{-0.15cm}
The generative model \eqref{eq:synthetic-data-model} leads to Gaussian class-conditional distributions with equal covariance matrices for both classes. Thus, Bayes-optimal classification can be achieved using a linear discriminant function $f^{\vec{w}}: \mathbb{R}^{D} \to \mathbb{R}$ parametrized by a weight vector $\vec{w}$. We here use two different implementations of linear logistic regression (LLR). The first one is part of scikit-learn~\citep{pedregosa_scikit-learn_2011}, where we use the default parameters, no regularization, and no intercept. This implementation is employed in combination with model-agnostic XAI methods (Pattern, PFI, EMR, FIRM, SHAP, Pearson Correlation, Anchors and LIME, see below). The second implementation is a single-layer neural network (NN) with two output neurons and softmax activation function, which was built in Keras with
a Tensorflow backend. The network is trained without regularization using the Adam optimizer. This implementation is employed for model-based XAI methods defined on neural networks only (Deep Taylor, LRP, PatternNet and PatternAttribution, see description below).
Note that, while both implementations should in theory lead to the same discriminant function, slight discrepancies in their weight vectors are observed in practice (see supplementary Figure~S8). 


\vspace{-0.4cm}
\subsection{XAI methods}\label{subsec:XAImethods}
\vspace{-0.15cm}
We assess the following model-agnostic and NN-based saliency methods. All methods can be used to generate global maps, while only some are capable of generating sample-based heat maps. All methods provide continuous-valued saliency maps $\vec{s}$, which are compared to the binary ground truth (encoded in the sets $\mathcal{F}_{\text{dep}}^{+}$ and $\mathcal{F}_{\text{dep}}^{-}$) using metrics from signal detection theory (see Section~\ref{subsec:performance}). Thus, we do not require any method to provide a dichotomization function $\vec{h}$.

\paragraph{Linear model weights (extraction filters)}

For linear models $f^\vec{w}(\vec{x}) = \vec{w}^\top \vec{x} + b$, the model weights $\vec{w}$ are most commonly used for interpretation. Thus, the function
\begin{align}\label{eq:linear_filters}
\vec{s}^{\text{filter}}(\mathcal{D}) & = \vec{w} 
\end{align}
provides a global saliency map, which we call the \emph{linear extraction filter}. Notably, saliency methods based on the gradient of the model output with respect to the input features \citep{baehrens_how_2010, simonyan2013deep} reduce to the extraction filter for linear prediction models \citep{kindermans_learning_2017}.

As has been noted in Section~\ref{sec:supppressors} and, in more depth, in \citet{haufe_interpretation_2014}, extraction filters are prone to highlight suppressor variables. This also holds for sparse weight vectors, as the inclusion of suppressor variables in the model may be necessary to achieve optimal performance \citep{haufe_interpretation_2014}.

\paragraph{Linear activation pattern}

The set $\mathcal{F}_{\text{dep}}^+$ can be estimated from empirical data by testing the dependency between the target $y$ and each feature $x_d$ using a statistical test for general non-linear associations \citep[e.g.,][]{gretton2007kernel}. For the linear generative model studied here, it is sufficient to evaluate the sample covariance $\Cov[x_d, y]$ between each input feature $x_d, d \in \{1, \ldots, D\}$ and the target variable $y$. To obtain a model-specific output, we can further replace the target variable $y$ by its model approximation $\vec{w}^\top \vec{x}$, leading to $s_d(\mathcal{D}) := \Cov[x_d, \vec{w}^\top \vec{x}]$, for $d \in \{1, \ldots, D\}$. The resulting global saliency map
\begin{align}\label{eq:linear_pattern}
\vec{s}^{\text{pattern}}(\mathcal{D}) & = \vec{S}_\vec{x} \vec{w}, 
\end{align}
where $\vec{S}_\vec{x} = \text{Cov}[\vec{x}, \vec{x}]$ is the sample data covariance matrix, called the \emph{linear activation pattern} \citep{haufe_interpretation_2014}. The linear activation pattern is a global saliency map for linear models that does not highlight spurious suppressor variables \citep{haufe_interpretation_2014}. Note that $\vec{s}^{\text{pattern}}$ is a consistent estimator for the coefficients $\vec{a}$ of the generative model in our suppressor variable example \eqref{eq:suppressor}. In particular, $s^{\text{pattern}}_2 \rightarrow 0$ for $N \rightarrow \infty$. 


\paragraph{Pearson correlation}
Since the linear activation pattern corresponds to the covariance of each feature with either the target variable or model output, a natural idea is to replace it with Pearson's correlation 
\begin{align}\label{eq:pearson}
s_d^{\text{corr}}(\mathcal{D}) := \Corr[x_d, \vec{w}^\top \vec{x}] = \frac{\Cov[x_d, \vec{w}^\top \vec{x}]}{\sqrt{\Var[x_d]\Var[\vec{w}^\top \vec{x}]}}
\end{align}
in order to obtain a normalized measure of feature importance. However, due to the normalization terms in the denominator, this measure is more strongly affected by noise than the covariance-based activation pattern. 

\paragraph{Permutation feature importance (PFI) and empirical model reliance (EMR)}\label{par:pfi}
The PFI approach was introduced by \citet{breiman_random_2001}
to assess the influence of features on the performance of random forests. 
The idea is to shuffle the values of one feature of interest, keep the remaining features fix,
and to observe the effect on the miss-classification rate. 
In a  same style, \citet{fisher_all_2019} introduced the notion of model reliance,
a framework for permutation feature importance approaches. 
In our work, we utilize the empirical model reliance (EMR), which measures the change of 
the loss function after shuffling the values of the feature of interest. As such, PFI and EMR provide global saliency maps.

\paragraph{Feature importance ranking measure (FIRM)}\label{par:firm}
The feature importance ranking measure \citep[FIRM,][]{zien_feature_2009} takes the underlying correlations of the features into account, by leveraging 
the conditional expectation of the model's output function, given the feature of interest, and
measuring its deviation. As such, FIRM provides a global saliency map. While intractable for arbitrary models and data distributions, FIRM admits a closed-form solution for linear models and Gaussian distributed data, which we implemented. Notably, under these assumptions, FIRM is equivalent to the linear activation pattern (see above) up to a re-scaling of each feature by its standard deviation \citep{haufe_interpretation_2014}.

\paragraph{Local interpretable model-agnostic explanations (LIME)}\label{par:lime} 
To generate a saliency map for a model's prediction on a single example,
LIME~\citep{ribeiro_why_2016} samples instances around that instance, and weights the samples according to their proximity to it.
LIME then learns a linear surrogate model in the vicinity of the instance of interest, trying to linearly approximate the local behavior of the model, which is then interpreted by examining the weight vector (extraction filter) of that linear model. As such, LIME inherits the conceptual drawbacks of methods directly interpreting gradients or model weights in the presence of suppressor variables.

\paragraph{Shapley additive explanations (SHAP)}\label{par:shap}
The Shapley value \citep{shapley1953value} is a game theoretic approach to measure the influence of a feature on the decisions of a model on a single example. Since its computation is intractable for most real-world settings, an approximation called SHAP~\citep{lundberg_unified_2017} has become widely popular. We use the linear SHAP method, including the option to account for correlations between features.

\paragraph{Anchors}\label{par:anchors}
Anchors~\citep{ribeiro_anchors_2018} seeks to identify a sparse set of important features for single instances, which lead to consistent predictions in the vicinity of the instance.  Features for which changes in value have almost no effect on the model's performance are considered unimportant.

\paragraph{Neural-network-specific methods}\label{par:function_based}
In addition to the model-agnostic XAI methods introduced above, a number of model-specific methods tailored to neural network architectures are considered. All these methods are based on modified backpropagation, but deal with nonlinearities in the network in a different way. For all methods, the implementation in the \texttt{innvestigate}\footnote{https://github.com/albermax/innvestigate}~\citep{alber2019innvestigate} package is used. 

\citet{simonyan_very_2015} proposed a sensitivity analysis, where pixels for which the model output is more affected by a shift in the input signal are considered more important. To this end, the gradient of the output with respect to the input signal is calculated.  However, as discussed above, the gradient of a linear model reduces to its model weights (extraction filters): $Grad_{\text{NN}} = w_{\text{NN}}$. DeConvNet~\citep{zeiler_visualizing_2014} and Guided Backpropagation~\citep{springenberg_striving_2015} are two additional methods that again reduce to the gradient/extraction filter for linear models. 

PatternNet \citep{kindermans_learning_2017} is conceptually similar to gradient analysis. However, rather than model weights, activation patterns are estimated per node and backpropagated through the network. For linear networks, PatternNet coincides with the linear activation pattern approach, although we observe slight deviations between the methods in practice.

Lastly, layer-wise relevance propagation \citep[LRP,][]{bach_pixel-wise_2015}, Deep Taylor Decomposition \citep[DTD,][]{montavon_explaining_2017}, and PatternAttribution~\citep{kindermans_learning_2017} aim to visualize how much the different dimensions of the input contribute to the output through the layers. As such, each node in the network is assigned a certain amount of `relevance', while keeping the total `relevance' per layer constant. For LRP, two different variants (`rules') are included: the $z$-rule and the $\alpha\beta$-rule
\citep{bach_pixel-wise_2015}.
Deep Taylor Decomposition (DTD) approximates the subfunctions learned by the different nodes by applying a Taylor decomposition around a root point and pooling the relevance over all neurons. Lastly, PatternAttribution \citep{kindermans_learning_2017} estimates the root point from the data based on the PatternNet approach. 

\vspace{-0.4cm}
\subsection{Measures of explanation performance}\label{subsec:performance}
\vspace{-0.15cm}

While numerous subjective criteria for evaluating the success of XAI methods have been proposed~\citep[e.g.,][]{schmidt_quantifying_2019, nguyen_quantitative_2020}, we here aim to provide objective, data-dependent, criteria using definition \eqref{eq:objective_importance}. Since, we know that statistical differences between classes are only present in features belonging to the set $\mathcal{F}_{\text{dep}}^{+}$, while features $\mathcal{F}_{\text{dep}}^{-}$ are entirely driven by non-class-specific, fluctuations, the dichotomization
\begin{align}
{h}^{\text{true}}_d = 
\left\{
\begin{array}{ll}
    1, & d \in \mathcal{F}_{\text{dep}}^{+} \\
    0, & d \in \mathcal{F}_{\text{dep}}^{-}
\end{array}
\right.
\end{align}
is used as a ground truth both for global and instance-based `explanations'. This binary ground truth is compared to the continuous-valued saliency map $\vec{s}(f^{\boldsymbol{\theta}}, \vec{x}^*, \mathcal{D}) \in \mathbb{R}^D$ of each XAI method. 

\emph{Explanation performance} is measured by comparing $\vec{h}^{\text{true}}$ with $\vec{s}$. To this end, saliency maps are rectified by taking the absolute value $|\vec{s}|$. As performance metrics, we use the area under receiver operating curve (AUROC) and the precision for a fixed specificity of 90\% (PREC90), which is obtained using the lowest threshold on $\vec{s}$ for which the specificity is greater or equal to 90\%. Results were similar when AUROC was replaced by average precision (see supplementary material). The PREC90 metric is based on the following consideration: while $\mathcal{F}_{\text{dep}}^{+}$ defines the set of features that any XAI method \emph{may} highlight, a particular machine learning model may actually use only a subset of them. Thus, we would like to penalize false negatives (features that are in $\mathcal{F}_{\text{dep}}^{+}$ but receive a low score according to $\vec{s}$) much less than false positives (features in $\mathcal{F}_{\text{dep}}^{-}$ that receive a high score). To this end, we evaluate the precision (fraction of truly important features among those estimated to be important) at a high decision threshold based on the consideration that good XAI methods should assign very high saliency scores only to truly important features. Truly important features receiving low scores thus do not influence this metric.

Performance metrics are evaluated per model for global XAI methods and per sample for instance-based XAI methods. In addition, instance-based rectified saliency maps are averaged to also yield global maps
\begin{align}
\vec{s}^{\text{global}}(f^{\boldsymbol{\theta}}, \mathcal{D}) = \frac{1}{N} \sum_{n=1}^N |\vec{s}^{\text{instance}}(f^{\boldsymbol{\theta}}, \vec{x}^n, \mathcal{D})| \;, 
\end{align}
the performance of which is also evaluated. Note that, in our setting, the input-output relationships between features and target are static. Thus, the same, global, ground-truth saliency map is expected to be reconstructed by each local explanation on average. While  individual explanations may be heavily corrupted by noise, this efect should be suppressed when averaging rectified heat maps across all samples, which is, therefore, considered a meaningful way to derive global explanations for instance-based XAI methods. Thus, instance-based XAI methods are evaluated both in terms of global and single-instance performance, while global XAI methods are only evaluated with respect to the former.

\vspace{-0.4cm}
\section{Experiments}\label{sec:experiments}
\vspace{-0.15cm}

\begin{figure}[ht]
    \begin{center}
        \centerline{\includegraphics[width=0.6\columnwidth]{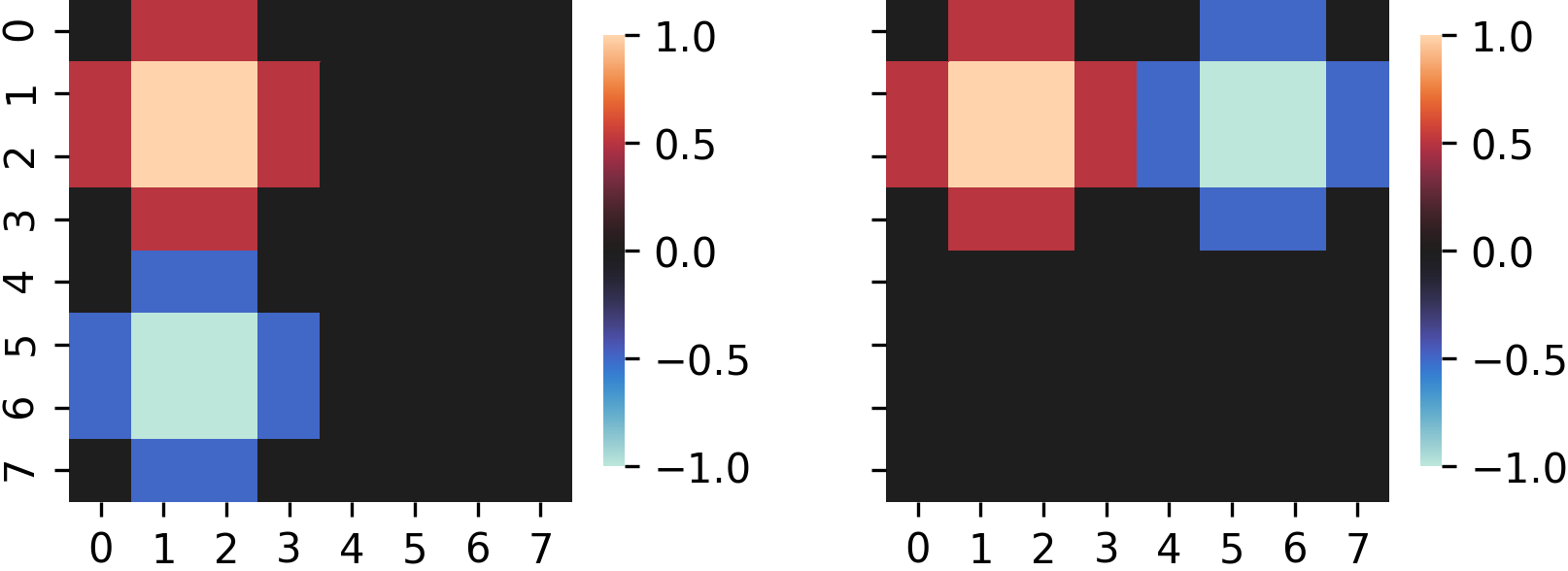}}
        \caption{The signal activation pattern $\vec{a}$ (left) and the distractor activation pattern $\vec{d}$ (right) used in our experiments can be visualized as images of 8 $\times$ 8 pixels size. The signal pattern consists of two blobs with opposite signs: one in the upper left and one in the lower left corner, while the distractor pattern consists of blobs in the upper left and upper right corners. Thus, the two components spatially overlap in the upper left corner.
        }
        \label{fig-pattern-matrix}
    \end{center}
    \vskip -0.4in
\end{figure}

We conduct a set of experiments aimed to address the following questions: 
(i) which XAI methods are best able to differentiate between important (that is, class-specific) features and non-important features,
(ii) how does the signal-to-noise ratio of the data (through the accuracy of the classifier) affect the explanation performance of each method. Python code to reproduce our experiments is provided on github\footnote{https://github.com/braindatalab/scrutinizing-xai}. We generate $K=100$ datasets with $N=1000$ samples each according to the model specified in Section~\ref{subsec:data-generation}. The prediction task is to discriminate between two categories encoded in the binary variable $y^n, n \in \{1, \ldots, N\}$. Our feature space are images of size 8 $\times$ 8, thus $D = 8^2 = 64$. Figure~\ref{fig-pattern-matrix} depicts the static signal and distractor patterns $\vec{a} \in \mathbb{R}^{64}$ and $\vec{d} \in \mathbb{R}^{64}$, which are identical across all experiments. As can be seen, signal and distractor overlap in the upper left corner of the image, while the lower left corner is occupied by the signal only and the upper right corner is occupied by the distractor only. All pixels are moreover affected by multivariate correlated Gaussian noise $\vecg{\eta}$. 

With the signal pattern $\vec{a}$, we control the statistical dependencies between features and classification target in our synthetic data. Therefore, the \emph{ground truth} set of important features in our experiments is given by
\begin{align}\label{eq:ground_truth_linear}
\mathcal{F}_{\text{dep}}^{+} = \{d \; | \; a_d \neq 0 \,, \; 1 \leq d \leq 64 \} \;.
\end{align}

Note that noise and distractor components both do not contain any class-specific information. The distractor, thus, merely serves as a strong one-dimensional noise component with predefined characteristic spatial pattern. Its main purpose in our experiments is to facilitate the visual assessment of saliency maps, where any importance assigned to the right half of the image represents a false positive. 

For each dataset, class labels $y^n$, distractor values $\rho^n$, and noise vectors $\vecg{\eta}^n$ are sampled independently from their respective distributions described in Section~\ref{subsec:data-generation}. Five different SNRs are analyzed, corresponding to five different choices of the parameter $\lambda_1 \in \{0.0, 0.02, 0.04, 0.06, 0.08\}$. Each resulting dataset $\mathcal{D}_{k, \lambda_1}, k \in \{1, \ldots, 100\}$ is divided into a train set $\mathcal{D}^{\text{train}}_{k, \lambda_1}$ and a validation set $\mathcal{D}^{\text{val}}_{k, \lambda_1}$, with  samples sizes $N^{\text{train}} = 800$ and $N^{\text{val}} = 200$, respectively. 

Linear logistic regression classifiers  $f^{\vec{w}}$ are fitted on $\mathcal{D}^{\text{train}}_{k, \lambda_1}$ and applied to $\mathcal{D}^{\text{val}}_{k, \lambda_1}$. The logistic regression implemented in scikit-learn is trained with a maximum number of 1000 iterations. The neural network based implementation is trained for 200 epochs with a learning rate of 0.1. Since the neural network has two output neurons, its effective extraction filter $\vec{w}^\text{NN}$ was calculated as the difference $\vec{w}^\text{NN} = \vec{w}^{\text{NN}}_1 - \vec{w}^{\text{NN}}_2$.

Saliency maps $\vec{s}(f^{\vec{w}}, \vec{x}^n, \mathcal{D}^{\text{train}}_{k, \lambda_1})$ are obtained for each of the methods described in Section~\ref{subsec:XAImethods} and for all datasets $\mathcal{D}^{\text{train}}_{k, \lambda_1}$, where instance-based maps are evaluated on all input examples $\vec{x}^n$ from the corresponding validation sets $\mathcal{D}^{\text{val}}_{k, \lambda_1}$. All XAI methods are applied using the default parameters, with the exception of $\text{LRP}_{\alpha\beta}$, for which we set $\alpha = 2$ and $\beta = 1$. 
That is, for SHAP, we use the \texttt{\small LinearExplainer} with \texttt{\small Impute Masker}, set \texttt{\small feature\_perturbation=correlation\_dependent} and perform the sampling with \texttt{\small samples = 1000}. For LIME, we use \texttt{\small kernel\_width} = $\sqrt{64} * 0.75$,
\texttt{\small kernel} = $\exp{(\nicefrac{-x^2}{\texttt{kernel\_width}^2})}^{\nicefrac{1}{2}}$, \texttt{\small discretize\_continuous = False}, and \texttt{\small feature\_selection = highest\_weights}. For ANCHORS, we use \texttt{\small threshold = 0.95},
\texttt{\small delta = 0.1},
\texttt{\small discretizer = 'quartile'},
\texttt{\small tau = 0.15},
\texttt{\small batch\_size = 100},
\texttt{\small coverage\_samples = 10000},
\texttt{\small beam\_size = 1},
\texttt{\small stop\_on\_first = False},
\texttt{\small max\_anchor\_size = 64},
\texttt{\small min\_samples\_start = 100},
\texttt{\small n\_covered\_ex = 10},
\texttt{\small binary\_cache\_size = 10000},
\texttt{\small cache\_margin = 1000}.

\vspace{-0.4cm}
\section{Results}
\begin{figure}[htb]
    \begin{center}
        \includegraphics[width=0.5\columnwidth]{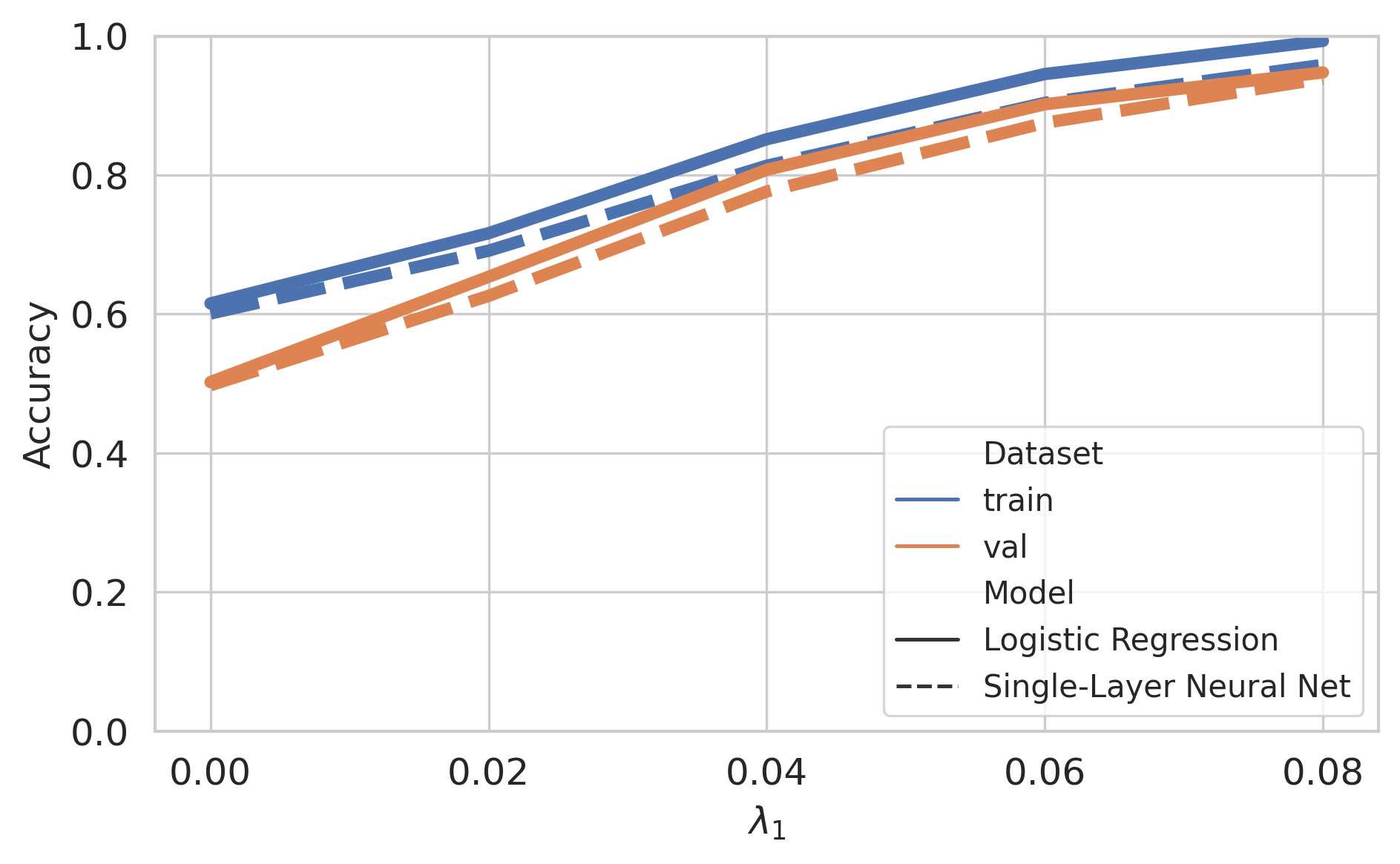}
        \vspace*{-0.1in}
        \caption{With increasing signal-to-noise ratio (determined through the parameter $\lambda_{1}$ of our generative model \eqref{eq:synthetic-data-model}), the classification accuracy of the logistic regression and the single-layer neural network increases, reaching near-perfect accuracy for  $\lambda_{1} = 0.08$.
        }
        \label{fig-avg-acc-plots}
    \end{center}
    \vskip -0.2in
\end{figure}

\vspace{-0.15cm}
Figure~\ref{fig-avg-acc-plots} shows the classification accuracy achieved by the LLR classifiers as a function of the SNR parameter $\lambda_1$. Both implementations reach near-perfect training and validation accuracy at an SNR of $\lambda_1 = 0.08$. For $\lambda_1 = 0$ (no class-specific information present), the validation accuracy attains chance level (0.5), as expected, while the training accuracy of 0.6 indicates a small degree of overfitting.

\vspace{-0.4cm}
\subsection{Qualitative assessment of saliency maps}\label{subsec:results-heat-maps}
\vspace{-0.15cm}

\begin{figure*}[htb]
    \begin{center}
        \includegraphics[height=3.1cm]{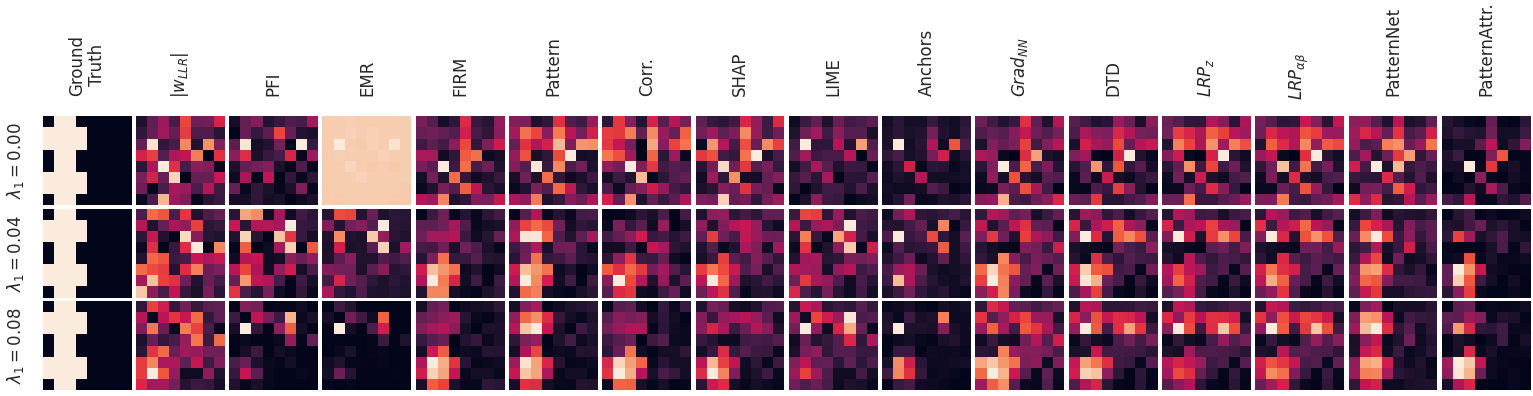}
        \caption{Global saliency maps obtained from various XAI methods on a single dataset. Rows represent three different choices of the SNR parameter $\lambda_1$. In the top row, no class-related information is present, yielding chance-level classification, while for the bottom row near-perfect classification accuracy is obtained. The `ground truth' set of important features is defined as the set of pixels with for which a statistical relationship to the class label is modeled, i.e. the set of pixels with nonzero signal patterns defined in \eqref{eq:ground_truth_linear}. Notably, a number of XAI methods assign significant importance to pixels in the right half of the image, which are statistically unrelated to the class label (suppressor variables) by construction.
        }
        \label{fig-heat-map-global-explanations}
    \end{center}
    \vskip -0.2in
\end{figure*}

\begin{figure}[htb]
    \begin{center}
        \includegraphics[height=3.1cm]{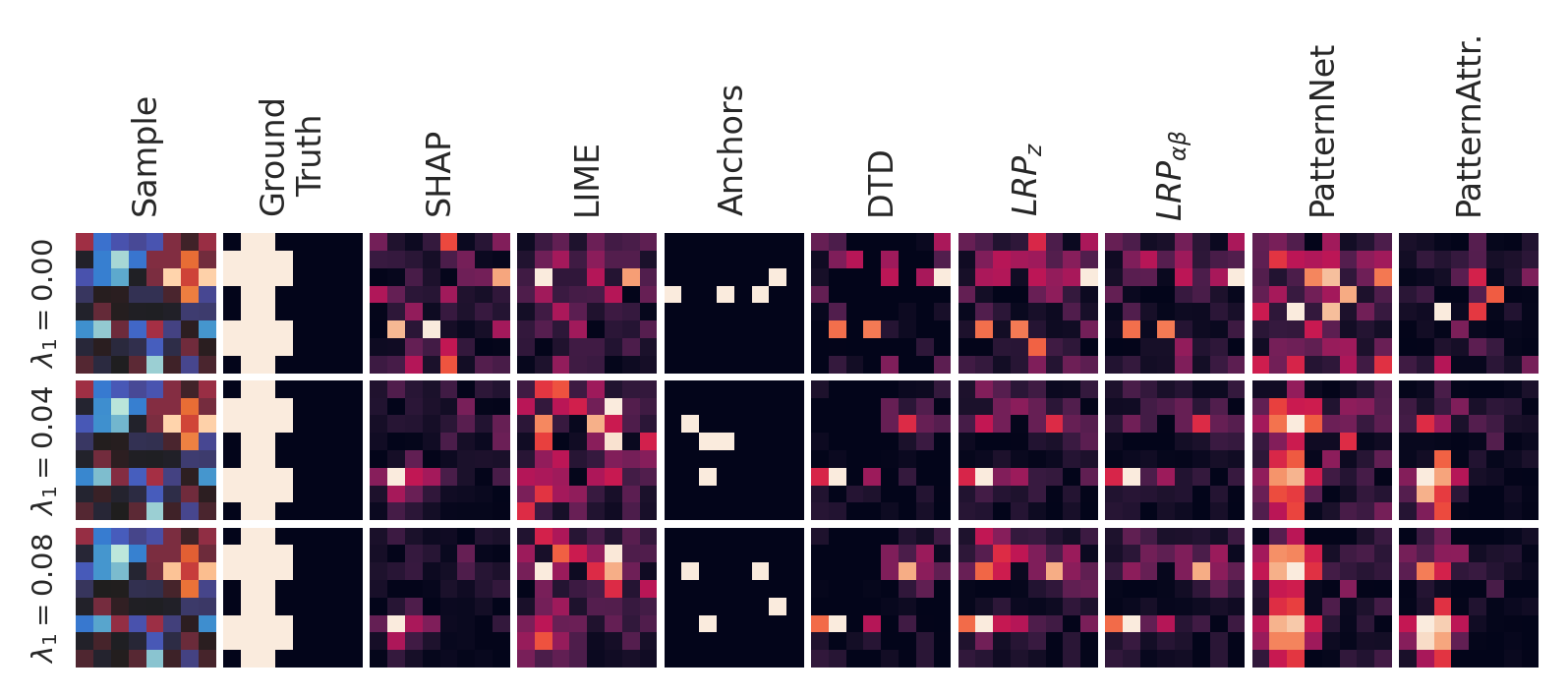}
        \caption{
        Saliency maps obtained for a randomly chosen single instance. At high SNR, PatternNet and PatternAttribution best reconstruct the ground truth signal pattern, while SHAP, LIME, DTD, and LRP assign importance to the right half of the image, where no statistical relation to the class label is present by construction,
        }
        \label{fig-heat-map-instance-explanations}
    \end{center}
    \vskip -0.2in
\end{figure}

Figure~\ref{fig-heat-map-global-explanations} depicts examples of global heat maps obtained for a randomly drawn dataset $\mathcal{D}_{k}$ for three different SNRs, $\lambda_1 \in \{0.0, 0.04, 0.08\}$. Shown are rectified quantities obtained by taking the absolute value. Instance-based maps were averaged over all instances of the validation dataset to obtain global maps. As expected, at $\lambda_1 = 0$ (lack of class-specific information; therefore, chance-level classification), none of the saliency maps resembles to the ground-truth importance map given by the pattern of the simulated signal. For  $\lambda_1 = 0.04$ and  $\lambda_1 = 0.08$, vast differences between different methods appear, though. The saliency maps of the linear Pattern as well as PatternNet deliver the best results on this dataset, recovering the two blobs of the ground-truth signal pattern most closely while correctly ignoring the right half of the image. FIRM, Correlation and PatternAttribution do recover the lower left blob of the signal pattern but assign much less importance to the upper left blob, where signal and distractor patterns overlap. All other methods including extraction filters $\vec{w}^{\text{LLR}}$ and $\text{Grad}_{\text{NN}}$, PFI, EMR, SHAP, LIME, Anchors, DTD, and the two LRP variants assign significant importance to the upper right corner, in which no class-related signal is present (thus, to suppressor variables), even for high SNR ($\lambda_1 = 0.08$). For some methods, the importance assigned to the distractor-only upper right blob is of the same order as the importance assigned to the upper left corner, in which signal and distractor overlap (PFI, EMR, LIME, Anchors, DTD and LRP). Interestingly, the signal-only lower left corner is assigned much less importance than the distractor-only upper right corner by some methods (PFI, EMR, LRP). This indicates that these methods mainly focus on the process of optimally \emph{extracting} the signal component with the distractor component rather than localizing the signal itself. Saliency maps provided by PFI, EMR, and Anchors are sparsest, focus on a small number of important as well as unimportant features, while gradient and extraction filter maps are the least sparse.


From the randomly chosen dataset $\mathcal{D}_k$ used to create Figure~\ref{fig-heat-map-global-explanations}, we further picked a single random instance that was correctly classified by both LLR implementations. Salience maps for this instance are shown in Figure~\ref{fig-heat-map-instance-explanations}. At high SNR, LIME, Anchors, DTD, and LRP still assign importance to the right half of the image, where no statistical relation to the class label is present by construction, while PatternNet, PatternAttribution, and SHAP do not. Interestingly, the instance-based saliency maps of the best performing methods, PatternNet and PatternAttribution, closely match the global maps obtained from these methods, even though they barely resemble features of the instance they were computed for. This suggests that these methods are strongly dominated by the global statistics of the training data rather than the properties of the individual input sample.  

\vspace{-0.4cm}
\subsection{Quantification of explanation performance}\label{subsec:results-quantification}
\vspace{-0.15cm}

\begin{figure}[htb]
    \begin{center}
        \includegraphics[height=6cm]{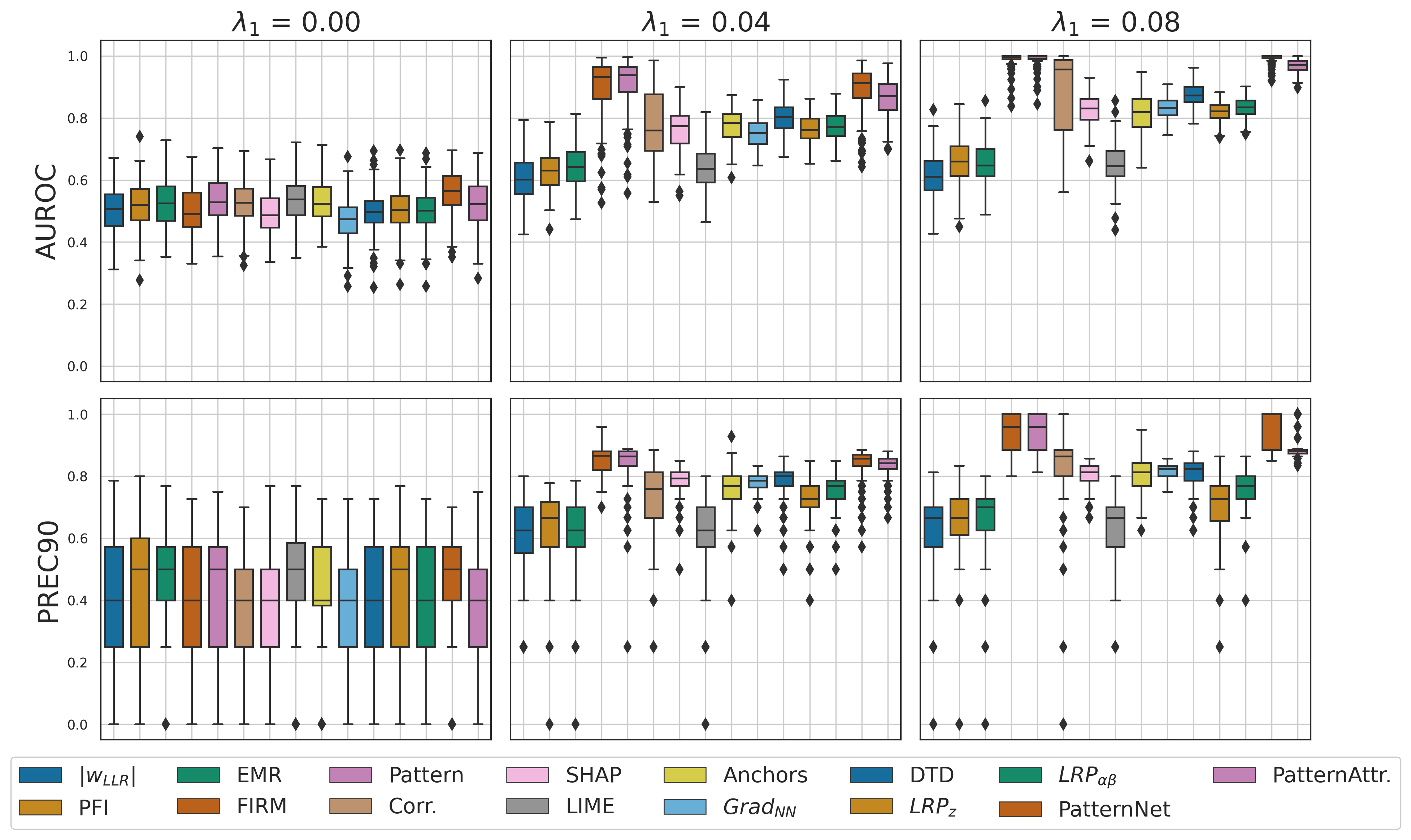}
        \caption{Quantitative explanation performance of global saliency maps attained by various XAI approaches. Performance was measured by the area under the receiver-operating curve (AUROC) and the precision at $\approx$ 90\% specificity. While chance-level performance is uniformly observed in absence of any class-related signal, stark differences between methods emerge for medium and high SNR ($\lambda_1 = 0.04$, and $\lambda_1 = 0.08$). Among the global XAI methods, the linear Pattern and FIRM consistently provide the best `explanations' according to both performance metrics. Among the instance-based methods, the saliency maps obtained by PatternNet and PatternAttribution (averaged across all instances of the validation set) show the strongest explanation performance.}
        \label{fig-box-plots-global}
    \end{center}
    \vskip -0.2in
\end{figure}

Figure~\ref{fig-box-plots-global} depicts the explanation performance of the global saliency maps provided by the considered XAI methods across 100 experiments. Shown are the median performance as well as the lower and upper quartiles, as well as outliers. As expected, the explanation performance of all methods (that is, the ability to distinguish truly important from unimportant pixels in the image) is not significantly different from chance level (AUROC = 0.5) when indeed no class-specific information is present ($\lambda_1=0$). At higher SNR ($\lambda_1 = 0.04$, and $\lambda_1 = 0.08$), most methods deviate from chance-level; however, significant differences are observed between methods. The median performances of LIME, PFI, and EMR quickly saturate at a low level around AUROC = 0.6. 
FIRM, Pattern, PatternNet and PatternAttribution have consistently
higher AUROC and PREC90 scores than other methods, approaching perfect performance for $\lambda_1 = 0.08$. The Pearson correlation between feature and class label can be considered as a runner-up, but is characterized by higher variance compared to the (covariance-based) Pattern. This can be explained by the fact that the presence of noise in the upper left corner (where both the signal and distractor are present) diminishes the correlation but not the covariance between the class label and the features in that corner. SHAP, DTD and LRP achieve moderate performance, while PFI, EMR, and Anchors do not perform well at all. This can only partially be explained by the sparsity of their saliency maps, which is penalized by the AUROC metric but not the PREC90 metric. Indeed, the difference between PFI, EMR, and Anchors on one hand and the rest of the methods on the other hand is smaller for the PREC90 than for the AUROC metric. However, the ranking of methods is similar for both metrics.

\begin{figure}[htb]
    \begin{center}
        \includegraphics[height=6cm]{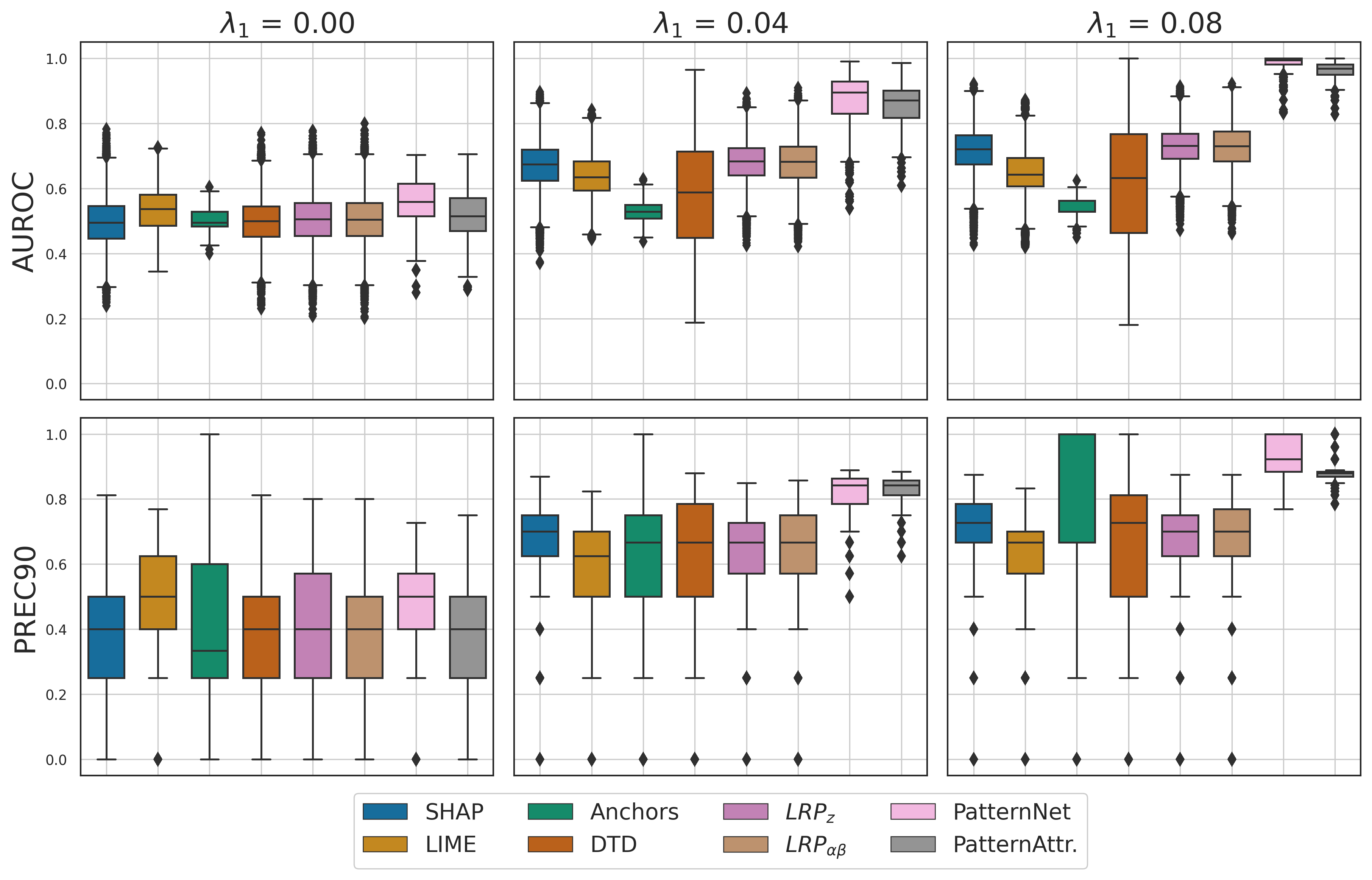}
        \caption{Explanation performance of instance-based saliency maps of neural-network-based XAI methods. All methods perform at chance level in absence of class-related information. For medium and high SNR, PatternNet and PatternAttribution show the strongest capability to assign importance to those feature that are associated with the classification target by construction, while ignoring suppressor features. }
        \label{fig-box-plots-sample-based}
    \end{center}
    \vskip -0.3in
\end{figure}

In Figure~\ref{fig-box-plots-sample-based}, quantitative results attained -- obtained per instance without averaging -- are shown. As observed for the global saliency maps, PatternNet and PatternAttribution the highest scores for both medium and high SNR, followed by the gradient of the neural network. Variants of LRP have achieve moderate performance in all settings. A high variance is, however, observed for DTD.

\vspace{-0.4cm}
\section{Discussion \& related work}\label{sec:discussion}
\vspace{-0.15cm}
`Explainable' artificial intelligence (XAI) is a highly relevant field that has already produced a vast body of literature. But many existing XAI approaches do not come with a theory on \emph{how} their results should be interpreted, i.e., what formal statements can be reasonably derived from them. We here formalize a minimal assumption that (as we believe) humans typically make when being offered `explanations'. Namely, that the input features highlighted by an XAI method must have an actual statistical relationship to the prediction target. Using empirical experiments and well-controlled synthetic data we demonstrate, however, that this is not guaranteed for a substantial number of state-of-the-art XAI approaches, inviting misinterpretations. In this light, interpretations such as those suggested for LIME in \citet{ribeiro_why_2016} (see introduction) seem to be unjustified, because LIME cannot rule out the influence of suppressor variables. False-positive associations between features and a disease thereby do not seem to be the only possible misinterpretations. A doctor confronted with the high importance of a variable known to be unrelated to a disease (a suppressor) may not be able to recognize that but may rather erroenously come to the conclusion that the model is not trustworthy. 

Our synthetic data were specifically designed to include suppressor variables, which are statistically independent of the prediction target but improve the prediction in combination with other variables. More specifically, suppressor variables display a \emph{conditional dependency} on the target given other features. In example~\eqref{eq:suppressor}, for example, the suppressor $x_2$ is independent of $y$ but becomes dependent on $y$ given $x_1$. A multivariate model can leverage this conditional dependency to improve its prediction -- here, by removing shared noise from feature $x_1$. However, `influential' features showing only such conditional dependencies can be of little interest in practice and need to be interpreted differently than features exhibiting a direct statistical relationship.

In our simulation, various XAI methods were found to be unable to reject suppressor variables as being unimportant. This failure was found to be aggravated in a setting where all signal-containing features were contaminated with the distractor (thus, the lower left blob in the signal pattern was absent), see supplementary material. While -- based on the consideration made above -- this behavior is expected for methods based on interpreting model weights, such as LIME or gradient-based approaches, it was also observed for PFI, EMR, SHAP, Anchors, DTD and LRP. While we suggest an explanation for that behavior in the following paragraph, future work will be required to theoretically study the behavior of each method in the presence of suppressors.        

The degree to which suppressor variables affect model explanations in practice is hard to estimate and may differ considerably between domains and applications. Thus, the quantitative results presented here are not claimed to universally hold. To rule out adverse effects to due suppressor variables or other detrimental data properties in a particular application, it should become common practice to conduct simulations with realistic domain-specific ground-truth data.

\vspace{-0.4cm}
\subsection{Insufficiency of model-driven XAI}
\vspace{-0.15cm}
In principle, one may argue that different types of interpretations can be useful in different contexts. For example, the identification of input dimensions that have a strong `influence' on a model's output may be useful to study the general behavior of that model (e.g. for debugging purposes). However, we argue that it is insufficient to analyze any model without taking into account the distribution of the data it was trained on. The difficulty of several XAI methods to reject suppressor variables can be explained by their inability to recognize that suppressor variables and truly target-related variables (e.g., $x_1$ and $x_2$ in example \eqref{eq:suppressor}) are \emph{correlated} and thus cannot be manipulated independently, limiting the degrees of freedom in which individual features can influence the model output. This limitation is not only inherent to several XAI methods but also to empirical `validation' schemes based on the manipulation of single input features. As the resulting surrogates do not follow the true distribution of the training data, limited insight about the actual behavior of the model when used for its intended purpose can be gained. 

In contrast to existing predominantly model-driven and data-agnostic XAI approaches, we here provide a definition of feature importance that is purely data-driven, namely the presence of a univariate statistical interaction to the prediction target. Importantly, this definition can also be tested on empirical data using statistical tests for non-linear interactions \citep{gretton2007kernel}. In the linear case studied here, it is sufficient analyze the covariance between features and prediction target, as described in \citep{haufe_interpretation_2014}, to obtain a saliency map with optimal explanation performance according to our metrics. The results of instance-based extensions of the linear covariance pattern, such as PatternNet and PatternAttribution \citep{kindermans_learning_2017}, however, suggest that the global covariance structure of the training may strongly dominate saliency maps obtained for single instances, which should be a subject of further investigation.

In general, features and target variables can be considered to be part of system of random variables whose relationships are governed by structural equations  (such as Eqs.~\eqref{eq:suppressor} and \eqref{eq:synthetic-data-model}). These structural relationships determine the possible statistical relationships of the involved random variables, and thus give rise to an even more fundamental definition of feature importance compared to our current definition based on actual statistical dependencies  (Eq.~\eqref{eq:objective_importance}). In fact, one can construct artificial settings, where structural relationships between features and target exist but do not manifest in statistical dependencies due to cancellation effects. However, we consider such situations rather irrelevant in practice. Typically, definitions based on structural and statistical relationship will coincide, which is also the case in our experimental setting. The advantage of definition \eqref{eq:objective_importance} is that, while structural relationships are hard to assess in practice, the mere presence of statistical relationships may be assessed empirically, offering a general way to estimate feature importance in practice.

Our definition encompasses the superset of features that \emph{may} be found important by any combination of machine learning model and saliency method. In fact, a model may not necessarily use all features contained in the set $\mathcal{F}_{\text{dep}}^{+}$ to achieve its prediction task. This is accounted for by our performance metric PREC90, which is designed to ignore most false negative omissions of important features. Practically, it may be desirable to \emph{fuse} data- and model-driven saliency maps, e.g. by taking the intersection between the estimated set $\mathcal{F}_{\text{dep}}^{+}$ and the set identified by a conventional XAI method.

\vspace{-0.4cm}
\subsection{Existing validation approaches}\label{sec:related-work}
\vspace{-0.15cm}
One can distinguish three categories of existing evaluation techniques for XAI methods. (i) evaluating the sensitivity or robustness of explanations to model modifications and input perturbations, (ii) using interdisciplinary and human-centered techniques to evaluate explanations, and (iii) establishing a controlled setting by leveraging a-priori knowledge about relevant features. 

\paragraph{Sensitivity-and robustness-centered evaluations}\label{para:sensitivity-eval}
Assessing the robustness and sensitivity of saliency maps in response to input perturbations and model changes is a common strategy underlying XAI approaches and their validation. However, such approaches do not establish a notion of correctness of explanations but merely formulate additional criteria (sanity checks)~\citep[see, e.g.,][]{doshi-velez_towards_2017}. For example, \citet{alvarez-melis_robustness_2018} assessed model `explanations' regarding their robustness -- asserting that similar inputs should lead to
similar explanations -- and showed that LIME and SHAP do not fulfill this requirement. Several studies developed tests to detect
inadequate explanations~\citep{adebayo_sanity_2018,ancona_unified_2017}. \citeauthor{adebayo_sanity_2018} demonstrated that, for some XAI methods, the identified features of trained models are akin to the ones identified by randomized models. \citet{hooker_benchmark_2019} came to similar conclusions. These features often represent low-level properties of the inputs, such as edges in images, which do not necessarily carry information about the prediction target \citep{adebayo_sanity_2018, sixt2020explanations}. 


\paragraph{Human-centered evaluations}\label{para:human-eval}
Human judgement is also often used to evaluate XAI methods~\citep[e.g.,][]{baehrens_how_2010,poursabzi2021manipulating,lage_human---loop_2018,schmidt_quantifying_2019}. To this end, the extent to which the use of model `explanations' can help a human to 
accomplish a task or to predict a model's behavior is typically measured. Another possibility is to define ground-truth explanations directly through human expert judgement \citep{Park_2018_CVPR}. As such, human-centered approaches also do not establish a mathematically sound ground-truth, as human evaluations can be highly biased. In contrast, we here exclusively focus on
formally-grounded evaluation techniques,  \citep[c.f.,][]{doshi-velez_towards_2017}.

\paragraph{Ground-truth-centered evaluations}\label{para:ground-tuth-eval}
Few works have attempted to use objective criteria and/or ground truth data to assess XAI methods. \citet{kim_interpretability_2018} used synthetic data to obtain qualitative `explanations', which were then evaluated by humans, while \citet{yang_benchmarking_2019} derived quantitative statements from synthetic data. However, in both cases, the ground-truth was not defined as a verifiable property of the data but as a `relative feature importance' representing how `important' a feature is to a model relatively to another model. In other works, the importance of features was defined through a generative process similar to ours~\citep{ismail_input-cell_2019,tjoa_quantifying_2020}. Yet, these works have not provided a formal, data-driven, definition of feature importance that would provide theoretical basis for their ground truth. Moreover, correlated noise settings leading to emergence of suppressor variables, have not been systematically studied in these works but have been shown to have a profound impact on the conclusions that can be drawn from XAI methods here.

\vspace{-0.4cm}
\subsection{Limitations and outlook}
\vspace{-0.15cm}
The present paper focuses on data with Gaussian class-conditional distributions with equal covariance, where linear machine learning models are Bayes-optimal. While this represents a well-controlled baseline setting, it is unlikely that solutions for the linear case transfer to general non-linear settings. Non-linear extensions of the activation pattern approach, such as PatternNet and PatternAttribution \citep{kindermans_learning_2017}, exist but have not been validated on ground-truth data. Our future work will address this gap by simulating non-linear suppressor variables, emerging through non-linear interactions between features. 

In non-linear settings, a clear distinction between features statistically related to the target or not may not always be possible. In tasks like image categorization, class-specific information might be contained in different features for each sample, depending on where in the image the target object is located. In extreme cases, objects of all classes may allocate the same locations leading to similar univariate marginal feature distributions whereas the discriminative information is contained in dependencies between features. Future work will be concerned with providing ground-truth definitions better reflecting this case.  

\vspace{-0.4cm}
\section{Conclusion}
\vspace{-0.15cm}
We have formalized feature importance in an objective, purely data-driven, way as the presence of a statistical dependency between feature and prediction target. We have further described suppressor variables as variables with no such statistical dependency that are, nevertheless, typically identified as important according to criteria that are prevalent in the XAI community. Based on linear ground-truth data, generated to reflect our definition of feature importance, we designed a quantitative benchmark including metrics of `explanation performance', using which we 
empirically demonstrated that many currently popular XAI methods perform poorly in the presence of so-called suppressor variables. Future work needs to further investigate non-linear cases and conceive well-defined notions of feature importance for specific non-linear settings. These should ultimately inform the development of novel XAI methods.

\vspace{-0.4cm}
\bibliographystyle{spbasic}
\enlargethispage{0.5cm}
\bibliography{
bib_all,
bibliography,
stefan}

\end{document}